\documentclass{article}

\usepackage{microtype}
\usepackage{graphicx}
\usepackage{subcaption}
\usepackage{booktabs} 
\usepackage{listings} 
\usepackage{hyperref}
\usepackage{alltt}

\usepackage{hyperref}
\newcommand{\head}[1]{\noindent \textbf{#1}}

\usepackage[accepted]{sysml2019}

\usepackage{xcolor}
\definecolor{bluekeywords}{rgb}{0.13,0.13,1}
\definecolor{greencomments}{rgb}{0,0.5,0}
\definecolor{greynumber}{rgb}{0.0,0.0,0.0}
\definecolor{redstrings}{rgb}{0.9,0,0}
\sysmltitlerunning{RLgraph: Modular Computation Graphs for Deep Reinforcement Learning}

\begin{document}

\twocolumn[
\sysmltitle{RLgraph: Modular Computation Graphs \\ for Deep Reinforcement Learning}



\sysmlsetsymbol{equal}{*}

\begin{sysmlauthorlist}
\sysmlauthor{Michael Schaarschmidt}{equal,cam}
\sysmlauthor{Sven Mika}{equal,rlcore}
\sysmlauthor{Kai Fricke}{hhsu}
\sysmlauthor{Eiko Yoneki}{cam}
\end{sysmlauthorlist}

\sysmlaffiliation{cam}{University of Cambridge}
\sysmlaffiliation{rlcore}{\textit{rlcore}}
\sysmlaffiliation{hhsu}{Helmut Schmidt University}

\sysmlcorrespondingauthor{Michael Schaarschmidt}{michael.schaarschmidt@cl.cam.ac.uk}

\sysmlkeywords{Machine Learning, Reinforcement Learning, SysML}

\vskip 0.3in

\begin{abstract}
Reinforcement learning (RL) tasks are challenging to implement, execute and test due to algorithmic instability, hyper-parameter sensitivity, and heterogeneous distributed communication patterns. We argue for the separation of logical component composition, backend graph definition, and distributed execution. To this end, we introduce RLgraph, a library for designing and executing reinforcement learning tasks in both static graph and define-by-run paradigms. The resulting implementations are robust, incrementally testable, and yield high performance across different deep learning frameworks and distributed backends.
\end{abstract}
]



\printAffiliationsAndNotice{\sysmlEqualContribution} 

\section{Introduction}
\label{introduction}
The recent wave of new research and applications in deep learning has been fueled by both hardware improvements and deep learning frameworks simplifying design and training of neural networks \cite{chen2015mxnet, abadi2016tensorflow, seide2016cntk, paszke2017automatic}. Reinforcement learning (RL) algorithms combined with deep neural networks have in parallel emerged as an active area of research due to promising results in complex control tasks \cite{Levine2016, Tobin2017, Silver2017}. However, their design and execution have not found similar standardization. This is a consequence of the highly varied resource requirements, scheduling, and communication patterns found in constantly evolving RL methods. Implementations hence require a high degree of customization. 

A number of RL libraries has emerged to focus on distinct aspects of managing such workloads. For example, OpenAI baselines provides reference implementations meant to reproduce specific benchmark environments (e.g. Atari) \cite{openaibaselines}. TensorForce provides a declarative API focusing on ease of use in applications \cite{Schaarschmidt2018}. Ray RLlib seeks to simplify distributing RL workloads by moving from hand-designed distributed communication to actor-based centralized execution on Ray \cite{Liang2018, ray}. 

While these libraries serve different purposes, many of them suffer from similar design problems leading to difficulties in testing, distributed execution, and extensibility. The root cause of these difficulties lies in a lack of separation of concerns. Composition of logical components defined within an RL algorithm is tightly coupled with code fragments specific to a deep learning framework (e.g. TensorFlow calls to define placeholders and variables). This leads to ill-defined APIs and also makes the reuse and testing of components difficult. Similarly, the often complex dataflow within RL algorithms is intertwined with control flow regulating (distributed) execution, environment interaction, and device management. This results in distributed execution and local device strategies being tied to specific algorithms. 

\begin{figure}[ht] 
\centering
\includegraphics[scale=0.7]{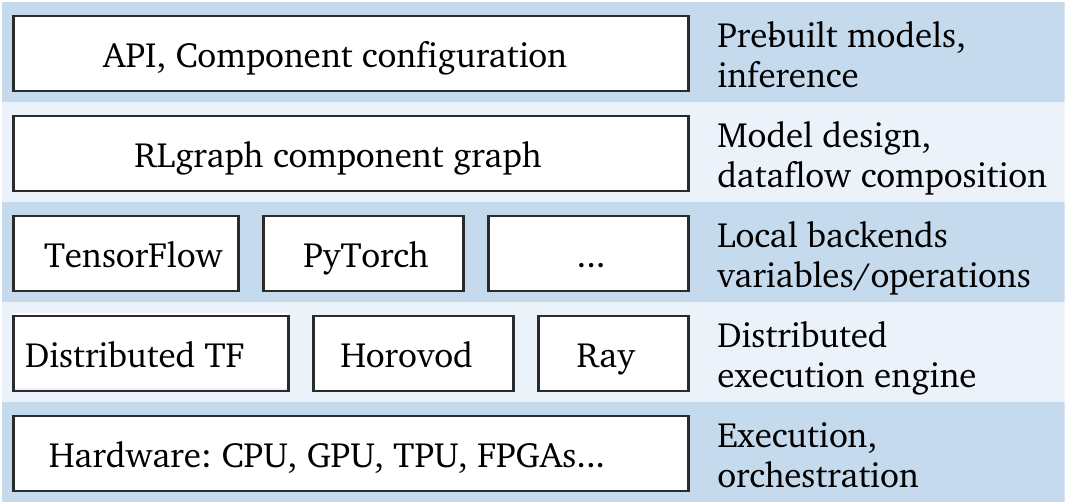}
\caption{RLgraph stack for using and designing RL algorithms.}
\label{fig:rlgraph-stack}
\vspace{-5mm}
\end{figure}

The central contribution of this paper is \textbf{RLgraph}, a modular framework to design and execute RL workloads from high-level dataflow. RLgraph addresses these issues by separating logical component composition, creation of operations, variables and placeholders, and finally local and distributed execution of the computation graph (Fig. \ref{fig:rlgraph-stack}). At the core of our design is a novel component graph architecture responsible for assembling and connecting algorithmic components, such as buffers or neural networks, and for exposing their functionality to a common API. Importantly, this component graph exists independently of implementation specific notions (e.g. TensorFlow variables), and instead relies on generalized space objects and operations. This means it can both be built for static graph and define-by-run backends, and RLgraph currently supports both TensorFlow (TF) and PyTorch (PT).

The component graph is built into a backend-dependent computation graph via a graph builder which generates operations, internal state (e.g. variables), device assignments, and a registry for the model's API. Developers are freed from tedious manual placeholder and variable definition, as they only need to specify type and shape of input spaces to an algorithm's outermost container component (root component). They can then rely on the RLgraph utilities to centrally handle most aspects of connecting complex models, e.g. splitting and merging complex nested spaces. A resource aware graph executor expands the component graph to add operations for local and distributed device strategies, e.g. by creating subgraph replicas for GPUs or managing globally shared state. At runtime, the graph executor (e.g. for TensorFlow) serves requests to the agent API by determining relevant input placeholders and operations from the op-registry and batching together all relevant operations into a single session call. Our design provides numerous advantages over many existing libraries:
\begin{enumerate}
\item \textbf{Distributed execution.} By separating concerns of design and execution, resulting agents can be implemented towards any distributed execution paradigm, e.g. using distributed TensorFlow \cite{abadi2016tensorflow}, Ray \cite{ray}, and plugins such as Uber's Horovod \cite{Sergeev2018}.
\item \textbf{Static and define-by-run backends.} The component graph does not impose restrictions on its execution. It supports end-to-end static graphs including control flow \cite{Yu2018} and define-by-run semantics (e.g. PyTorch \cite{paszke2017automatic}) through a unified execution interface.
\item \textbf{Fast development cycles.}  RLgraph's abstractions enables users to focus on high level dataflow when composing components. The build process manages backend scaffolding and creates operations based on user-provided input spaces.
\item \textbf{Incremental building and testing.} Existing libraries cannot efficiently identify problems in individual components as they do not offer a modular build system. In RLgraph, all components (including pre/post-processing heuristics) are first-class citizens  which are individually built and incrementally tested.
\end{enumerate}

In the remainder of the paper, we analyze RL design problems and survey existing libraries (\S \ref{motivation}). We then discuss the design of RLgraph (\S \ref{framework-design}, \S\ref{execution}). In the evaluation, we compare RLgraph against reference implementations using different execution paradigms (\S \ref{evaluation}). Our results show RLgraph can improve sample throughput over existing implementations by up to 180\%. In related work, we discuss emerging approaches in programming models and optimization (\S \ref{related}). RLgraph is available as open source\footnote{\url{https://github.com/rlgraph/rlgraph}}.

\section{Motivation}
\label{motivation}
\subsection{RL workloads}\label{workloads}
The central difficulty of executing RL workloads lies in the need for frequent interaction with the problem environment during training to evaluate and update the model. Environments may take the form of expensive physical systems (robots), 3D scene simulators, games, or generally any system exposing a state representation and an action interface. This is in contrast to supervised workloads, where training data is typically entirely available in advance, thus enabling straightforward batching and synchronization strategies \cite{Sergeev2018}. As a consequence of fast moving and empirically driven research, RL algorithms vary across all dimensions of execution (recently discussed by Liang et al. \cite{Liang2018}).

\head{State management.} Sample trajectories are often collected in a distributed fashion where workers interact with dedicated (simulation) environment copies. Algorithms manage synchronization of model weights between one or multiple learners and sample collectors, employing synchronous and asynchronous strategies. In addition, they must process and transmit samples to learners efficiently, sometimes involving hierarchies of local and distributed shared buffers to split post-processing tasks \cite{Horgan2018}.

\head{Resource requirements and scale.} Recent successes in applying RL at scale in gaming (e.g. OpenAI Five \cite{openaidota}, AlphaGo \cite{Silver2017}) were enabled by training models on up to tens of thousands of CPU cores and hundreds of GPUs. In contrast, models for environments which are not easily parallelized may be executed on a single CPU but might have stringent latency requirements.

\head{Models and optimization strategies.} Neural networks used to represent policies range from small multi-layer perceptrons to complex hierarchical representations \cite{SilverHuangMaddisonEtAl2016,Wayne2018}. Learning approaches vary from small incremental updates to expensive but infrequent policy optimizations over large batches, making effective use of hardware accelerators difficult. 

\subsection{Existing abstractions}\label{existing-libraries}
\head{Reference implementations.} Many libraries primarily serve as reference implementations. For example, \textbf{OpenAI baselines} \cite{openaibaselines}, \textbf{Keras-rl} \cite{plappert2016kerasrl} and Google's \textbf{Dopamine} \cite{dopamine} provide collections of well-tuned algorithms on benchmarks such as OpenAI gym \cite{openaigym} or ALE \cite{Bellemare2013}. \textbf{Nervana Coach} \cite{nervanacoach} contains a similar collection but with added tools for visualizing progress, and facilities for hierarchical learning and distributed training. \textbf{Horizon} focuses on building end-to-end pipelines for off-policy training at Facebook \cite{Gauci2018}.

Reference implementations share some components between algorithms (e.g. network architectures) but typically ignore many practical considerations in favour of concise code. Retooling them to different execution modes, environment semantics, or device strategies (e.g. multi-GPU support) requires significant work due to hard-coded, tightly coupled designs. 

\head{Centralized control.} \textbf{Ray RLlib} \cite{Liang2018} defines a set of abstractions for scalable RL. It relies on Ray's actor model \cite{ray} to execute RL algorithms via centralized control. At the core of RLlib's hierarchical task parallelism approach lies a set of optimizer classes. Each optimizer implements a \textit{step()} function which distributes sampling to remote actors, manages buffers, and updates weights. For example, an \textit{AsyncReplayOptimizer} implements distributed prioritized experience replay \cite{Horgan2018}.  Each step, the optimizer loop pulls samples from actors, inserts them into replay buffers, and performs training on an asynchronous learner thread. A core claim of RLlib is the separation of the execution plane in the optimizer from the definition of the RL algorithm within a policy graph. However, each optimizer encapsulates both local and distributed device execution. This means for example that only the dedicated multi-gpu optimizer class supports splitting input batches synchronously over multiple GPUs.  RLlib's optimizer abstractions also mix Python control flow, Ray calls, and TensorFlow calls throughout its components. Algorithms implemented in RLlib are hence not easily portable as training is principally meant to be executed only on Ray. 

\head{Fixed end-to-end graphs.} \textbf{TensorForce} \cite{Schaarschmidt2018} is a TF library providing a declarative interface to a number of RL algorithms. TensorForce focuses on applied use cases where control flow is driven by external application contexts, not simulation environments. Its end-to-end in-graph control flow design accelerates execution by avoiding unneeded context switches between Python interpreter and TF runtime \cite{Yu2018}. A key disadvantage of this design (also adopted by Batch PPO and TF-Agents \cite{Hafner2017,TFAgents}) is that partial data-flow is difficult to test due to limited in-graph-debugging facilities. Further, execution assignments via device and variable sharing decorators are not separate from algorithm logic in the absence of a modular build process.

\section{Framework design}\label{framework-design}
\subsection{Design principles}
In the absence of a single dominant design pattern, frameworks must resolve the tension between flexible prototyping, reusable components, and scalable execution mechanisms. RLgraph's design is driven by a number of insights:

\head{Separating algorithms and execution.} RL algorithms require complex control flow to coordinate distributed state and sample collection on one hand, and internal training logic on the other hand. Separating these aspects is difficult but essential to avoid re-implementing execution strategies. RLgraph manages local execution via graph executors which expose a clear interface between a high level API and the component graph. Distributed coordination is delegated to dedicated distributed executors (e.g. on Ray), or as part of the graph build in the executor (distributed TF).

\head{Reusable components with strict interfaces.} Deep learning frameworks enable quick prototyping of neural networks by exposing APIs to combine different types of layers with compatible interfaces. Providing a similar set of interchangeable components towards RL is complicated by the multitude of learning and execution semantics. This is exacerbated by implementations containing definitions in multiple execution contexts, e.g. Python control flow interleaved with calls to TF runtime. Tight coupling of components, and in turn a lack of well-defined interfaces and component boundaries, means that re-usability is severely constrained. RLgraph addresses this problem via its modular component architecture. Components only interact via declared API methods and guarantee reusability as they are fully specified through compatible input spaces.

\head{Incremental sub-graph testing.} An undesirable consequence of incorporating stochastic approximations at all levels is numerical sensitivity and non-determinism \cite{Nagarajan2018}. RL algorithms can require an overwhelming number of hyperparameters (often in excess of 25). This has created severe issues for robustness and reproducibility~\cite{Henderson2017,Mania2018}. Implementations are notoriously difficult to debug and test in part because generating and verifying inputs and outputs of partial dataflow is tedious (e.g. manually creating tensors of required shapes). RLgraph enables sub-graph testing by allowing users to send example data from input spaces through arbitrary components and component combinations.

\subsection{Components and graphs}
\head{Components.} Next, we discuss the design of RLgraph's component graph. For simplicity, we use TensorFlow as the primary backend and describe the implementation of other backends (e.g. PyTorch) in \S \ref{backends}. RLgraph's core abstraction is the \textit{Component} class which encapsulates arbitrary computations via \textit{graph functions}. RLgraph components are conceptually similar to DeepMind Sonnet's components \cite{sonnet} but offer more advanced notions of composition.
\begin{figure}[h] 
\centering
\includegraphics[scale=.4]{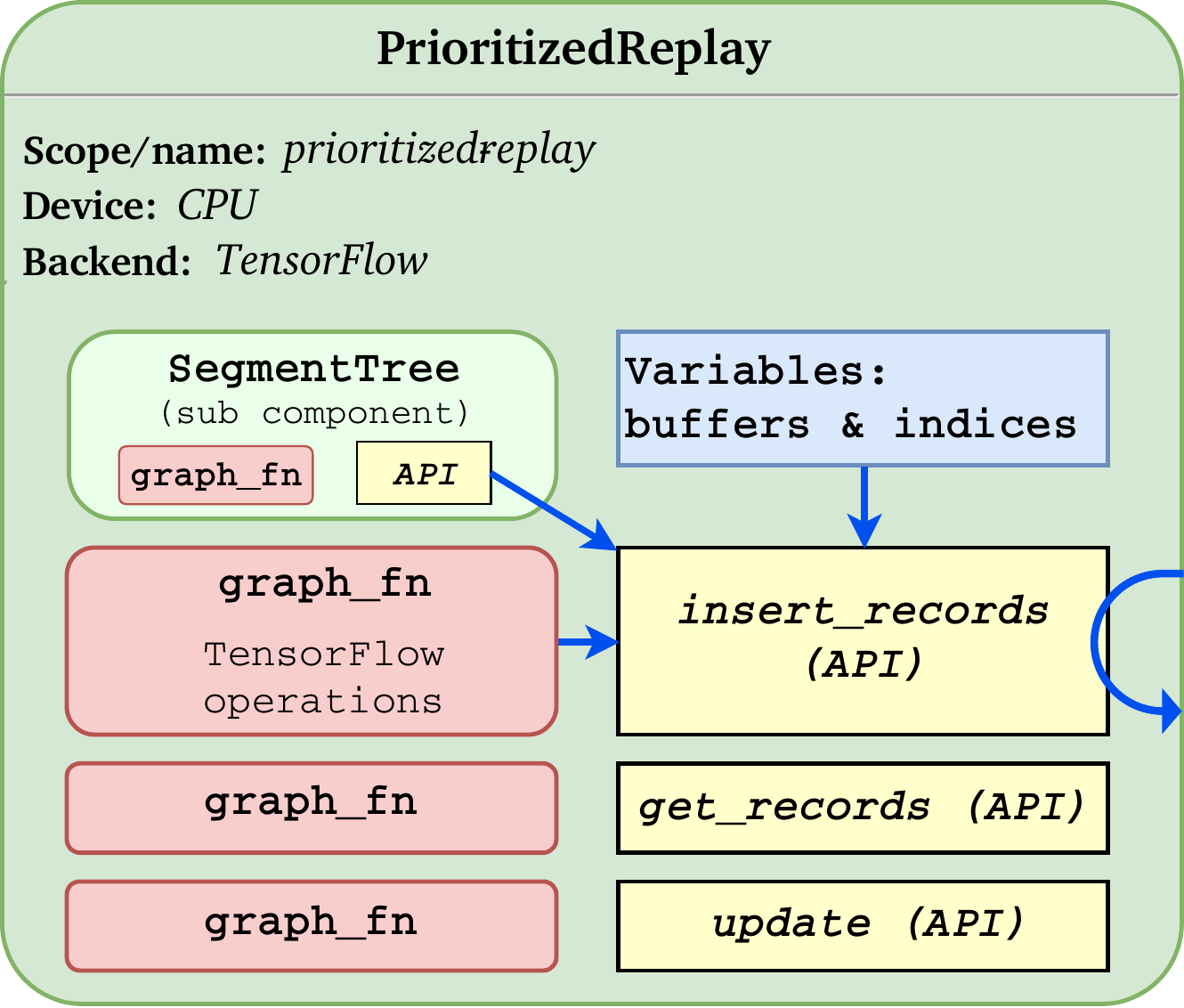}
\caption{Example memory component with three API methods.}
\label{fig:example-component}
\vspace{-2mm}
\end{figure} 
Consider a replay buffer component which exposes functionality to insert experiences and sample mini-batches according to priority weights. Implementing this buffer in an imperative language such as Python is straight-forward, but including it as part of a TensorFlow graph requires creating and managing many variables through control flow operators (e.g. to update priorities). Composing multiple such components in a re-usable way is difficult due to an impedance mismatch between class-based programming in a driver language, and functional transformations within a dataflow graph. Using a define-by-run framework (e.g. PyTorch) eases development but can create difficulties in large scale distributed execution and program export.

Existing high-level APIs for neural networks such as Sonnet, \cite{sonnet}, Keras \cite{chollet2015keras}, Gluon \cite{gluon}, or TF.Learn \cite{Tang2016} focus on assembly and training of neural networks. Implementing RL workloads in these frameworks usually means mixing imperative Python control flow with deep learning graph objects, leading to the design issues discussed before. 

When building for a static graph backend, RLgraph's component API enables fast composition of end-to-end differentiable dataflow graphs with in-graph control flow. The graph builder and executor automatically manage burdensome tasks such as variable and placeholder creation, scopes, input spaces, and device assignments. 

\head{Example component.}
In figure \ref{fig:example-component}, we show a simplified prioritized replay buffer component. All components inherit from the generic \textit{Component} class and assemble logic by combining their own sub-components. The buffer has a segment tree sub-component to manage priority orders. It exposes API methods to insert, sample, and update priorities which under the hood map to three graph functions. The difference between simple object methods and RLgraph API methods is that registered API methods are identified and managed in the build. Input shapes can be inferred automatically via dataflow from inputs to the root component.

Developers can declare methods as API methods by calling a register function (or in the future using a decorator). Technically, not all functionality of a component needs to be registered as an API method. Users can also implement helper functions or utilities, e.g. using TensorFlow operations without including them as API methods, if they do not need to be called from external components. Implementing such utilities as RLgraph components with API methods is nonetheless useful because these features can then be built and tested as sub-graphs.

Components can call arbitrary sub-components (and their sub-components). A component may have multiple API methods where input spaces to one method depend on outputs of its other methods. The build process ensures component computations and internal variables are only created once its input spaces are known. Instead of creating implicit device assignments, scopes, and variable sharing through nested contexts, RLgraph explicitly manages these properties per component.

\subsection{Building component graphs}
RLgraph models are assembled in three distinct phases:
\begin{enumerate}
\item \textbf{Component composition phase} in which component objects are defined and combined, including arbitrary nesting of sub-components. 
\item \textbf{Assembly phase} in which a type- and dimension-less dataflow graph (the component graph) is created. This is achieved by
calling each of the root component's API methods once to traverse its graph. The API-methods of the root component define the externally visible API of the component graph.
\item \textbf{Graph compilation/building phase} in which all computation operations are defined for each component. Inside a component class, these definitions are placed in special functions called \textit{graph functions}. Graph functions are the only places in the code where backend dependent objects are used (e.g. TF ops).
\end{enumerate}

There are also intermittent sub-phases for the initialization of execution aspects (e.g. session management) which are explained  in \S \ref{execution}.
A wide range of off-the-shelf component implementations such as buffers, optimizers, neural networks, or nested space splitters and mergers means that most users will only need to define few components to prototype new algorithms (e.g. loss function, network architecture).

\begin{figure}[t] 
\hspace{-2mm}
\includegraphics[scale=.65]{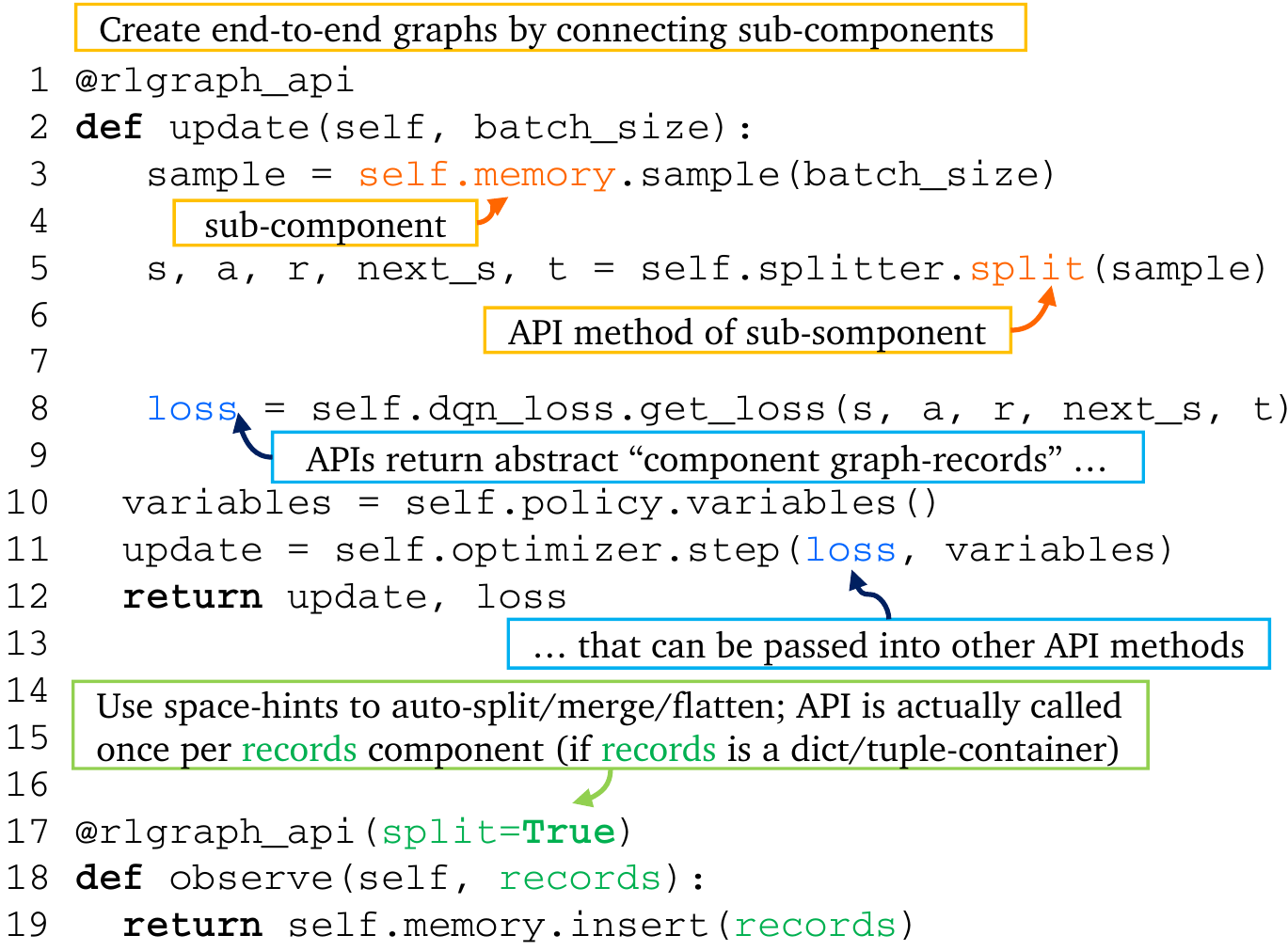}
\vspace{-6mm}
\caption{Example API methods show framework-independent control-flow construction.}
\label{fig:rlgraph-api-example}
\vspace{-6mm}
\end{figure} 

\head{1. Component composition and nesting.}
All components needed by a model are defined as Python objects. Components are logically organized as nested subcomponents of a root container component exposes the external API. Note that an agent can in theory define multiple root components
to act and learn different policies in parallel.

\head{2. Assembling the component graph.}
Next, users define the dataflow through the model via connections between components. Neither data types nor shape information are necessary at this stage. Each
component comes with a set of API-methods (Figure \ref{fig:rlgraph-api-example}).
The data (tensors) are interpreted inside these methods as abstract meta-graph operator objects, and their shapes and types will be inferred at build time.  This is achieved via decorators for API and graph functions which provide options to split, merge, un-nest/re-nest inputs and outputs. Return values of an API call can now be passed into other API-methods, a sub-component's API-method, or into a graph function for numerical manipulation. RLgraph spaces can conveniently nest, merge, split and fold time and batch dimensions of tensors through components. In our experience, these utilities drastically reduce development times as the different build phases automatically detect problems when manipulating complex spaces, e.g. records containing multiple states and actions with batch and time dimensions.

A simplified component graph assembly procedure is shown in Algorithm \ref{meta-graph-build}. The root component exposing the external interface (e.g. \textit{act}, \textit{observe}, \textit{update}) and the input spaces for the external API are passed to the component graph builder. This builder generates the backend-independent dataflow graph and the API by iterating over all API methods defined in the root component. For each method, a component graph op is created for each of its parameters and looked up in the input graph (type checks and default argument handling omitted). The component graph is then traversed through composed API functions which infer parameters and return values for each call , and these are stored as records in the component graph. Finally, the API method is registered in an API registry which contains the input spaces and final output ops (identified through the traversal).

\begin{algorithm}[t!]
   \caption{Component graph build procedure\label{meta-graph-build}}
\begin{algorithmic}
\STATE {\bfseries Input:} component $root$, input\_spaces $spaces$
\STATE $api=dict()$
\STATE // Call all api methods once, generate op columns.
\FOR{$method, record$ {\bfseries in} $root.api$}
\STATE $in\_ops\_records = list()$
\STATE // Create one input record per API input param.
\FOR{$param$ {\bfseries in} $record.input\_args$}
\STATE $in\_ops\_records.append(Op(param.space)) $
\ENDFOR
\STATE // Traverse graph from root for this method.
\STATE $out\_ops\_records = method(in\_ops\_records)$
\STATE // Register method with graph inputs and output ops.
\STATE $api[method] = [in\_ops\_records, out\_ops\_records]$
\ENDFOR

\STATE {\bfseries return} ComponentGraph(root, api)
\end{algorithmic}
\end{algorithm}

\head{3. Building computation graphs.}
\label{graph-build}
All numerical operations occur in the third phase inside graph functions which implement each component's API. Operations defined in a graph function include for example defining loss functions, or sampling from a buffer. As sometimes the component's variables must be accessed to complete these operations (e.g. a neural network layer must read its weights), RLgraph ensures that any such computation function is only called after all variables of a component have been defined. For example, the memory component in Fig. \ref{fig:example-component} can only define its buffers (e.g. TensorFlow variables) once it receives shapes and types of buffer contents. 

This barrier is enforced during the build, and custom components only need to override a generic method for variable creation. The method is called automatically and receives types and shapes of variables as input arguments. Developers thus only need to specify the external input spaces to the program (e.g. int/float boxes with batch and time ranks, container spaces for nested data). In practice, these are primarily state and action layouts defined by the environment.

We briefly describe the intuition behind the main build algorithm. The build begins by calling all API methods defined at the root from the provided input spaces until a component is input-complete, i.e. all spaces for all its computations are available. It then executes a completion function which calls the component's \textit{create\_variables}, and subsequently its graph functions under the correct device and scope to define its operations. Once a component is complete, the outputs of its graph functions become available as input spaces for subsequent components/graph functions. We then simply perform breadth-first-search until there are no more components to build or a constraint violation is detected. Input placeholders and op names are created and stored for all ops and output combinations defined in the API. 

\begin{lstlisting}[caption={Testing sub-graphs from arbitrary spaces.},aboveskip=3mm,label={li:test-example},moredelim={[is][emphstyle]{@@}{@@}}]
state_space = FloatBox(shape=(64,), add_batch_rank=True,
		 	add_time_rank=True)
# Dict space: 1 discrete, 1 continuous action.
action_space = Dict(discrete=IntBox(),
			cont=FloatBox(), add_batch_rank=True)
policy = Policy("recurrent_policy.json", action_space)

# Construct sub graph from spaces, auto-gen placeholders.
test = ComponentTest(policy, dict(nn_input=state_space),
	action_space=action_space))
# Test with any inputs in the input space.
action = test.test(policy.get_action, state_space.sample())
\end{lstlisting}

\head{Testing sub-graphs.} Identifying bugs in sub-tasks is difficult without a systematic mechanism for fine-granular input generation. Consider the RLgraph test class example in Listing \ref{li:test-example}. Here, we build a \textit{Policy} component (with subcomponents for a recurrent network and action selection) for the specified state and action spaces (with options for batch and time ranks). The test helper builds the sub-graph for the policy via the phases describe above. Users can then run the test and call an API method (e.g. by sampling an input from the input space). This call is delegated to a graph executor which executes the corresponding op. Every component (including all pre/post processing and learning heuristics) is continuously tested separately and in integration tests for larger graphs (i.e. complete RL algorithms). 

\subsection{Agent API}
Pre-built models can be configured via declarative configurations similar to TensorForce \cite{Schaarschmidt2018}. Configurations are provided as e.g. JSON documents specifying an algorithm and its components (network with list of layers, buffer, optimizers, device strategy etc.). The agent interface defines a set of abstract methods which agents must support to access certain execution modes (Listing \ref{li:agent-api}). 

The main difference between RLgraph and existing APIs lies in strictly enforced component boundaries and more explicit execution semantics. Fine-grained device control is managed via a device map where each components operations and variables can be assigned separately and selectively. Components may only exchange data along edges of the component graph where an edge corresponds to a call to a declared API method. This ensures well-defined APIs and also avoids a common case where two components are implemented to always be used together, thus making individual reuse difficult. In RLgraph, all components can be used and built individually from any input spaces. Next, we discuss graph execution.

\begin{lstlisting}[caption={High level agent API.},belowskip=-4mm,label={li:agent-api},moredelim={[is][emphstyle]{@@}{@@}}]
abstract class rlgraph.agent:
  # Build with default devices, variable sharing, ..
  def build(options)
  def get_actions(states, explore=True, preprocess=True) 
  # Update from internal buffer or external data.
  def update(batch=None, sequence_indices=None)
  # Observe samples for named environments.
  def observe(state, action, reward, terminal, env_id)
  def get_weights, def set_weights
  def import_model, def export_model
\end{lstlisting}

\section{Executing graphs}\label{execution}
\subsection{Graph executors.}
All build phases are invoked from a \textit{graph executor} which serves as the execution bridge between the component graph and a backend framework. Graph executors expose an \textit{execute()} method which takes name and arguments of an API method and returns the result. They further manage any backend-specific initialization, monitoring, and devices. For example, the TensorFlow executor assumes the following tasks. First, it initializes the TensorFlow session and variables, and builds hooks for summaries or profiling. For operation execution, it fetches input placeholders and op names from the graph operation registry and assembles session inputs. Importantly, there is no other interaction between user programs and graph other than through API operations defined in the  root component.

\head{Device management.}
Graph executors also handle device assignments by interleaving the build with a phase to initialize device strategies. Upon initialization, local device information (e.g. CUDA visible devices) is read and compared against user-defined device maps or synchronization modes. Consider a synchronous multi-GPU strategy where each input batch is split to one graph copy per GPU, with gradients averaged for the final update. Managing this synchronization requires the creation of additional operations and variables for the device copies and the splitting and averaging logic. The graph executor does this by creating core component copies and connecting them through generic input space splitters before building the component graph. Since no variables or placeholders have been created at this stage, components or sub-graphs can straightforwardly be modified, replaced or extended. In addition to generic device strategies, users can define a device map which specifies a device assignment for each component's ops and variables.

\head{Distributed execution.}
RLgraph can be executed in distributed mode either with framework-specific mechanisms or by using any other means to create and synchronize agents. For example, when using distributed TensorFlow, the TF graph executor can create the necessary parameter server and global/local synchronization operations. It can also plug-in third party tools such as Uber's Horovod to assume specific aspects of distributed communication for a backend (e.g. ring all-reduce \cite{Sergeev2018}). 
\begin{figure}[t] 
\centering
\includegraphics[scale=.55]{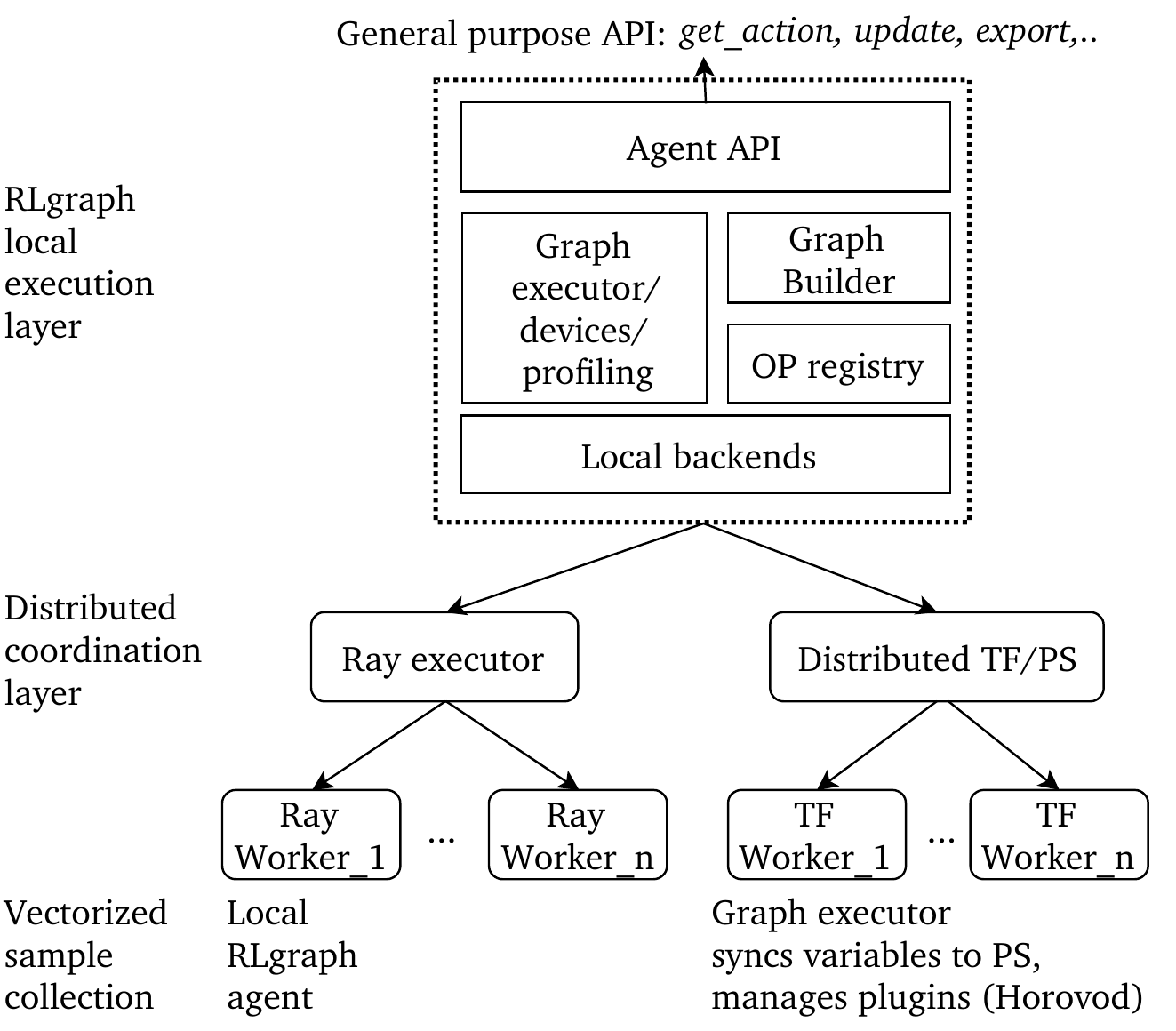}
\vspace{-2mm}
\caption{RLgraph execution stack.}
\label{fig:rlgraph-graph-execution}
\vspace{-6mm}
\end{figure} 
To demonstrate RLgraph's flexibility, we also built a Ray executor which can execute arbitrary RLgraph implementations on Ray's centralized execution model \cite{ray}. In the evaluation, we show that our implementation outperforms Ray's native library RLlib \cite{Liang2018}. 

Figure \ref{fig:rlgraph-graph-execution} illustrates how local execution and distributed coordination are separated. When using Ray, we simply pass an agent configuration to our Ray executor which creates Ray workers, each locally generating their component graph and graph executor to interact with environments. When using non-centralized control (e.g. distributed TensorFlow), the graph executor creates the necessary session and server objects and handles parameter server synchronization on each worker. Common features such as fault tolerance are delegated to the underlying execution engines via a configuration interpreted by the respective graph executors.

\subsection{Backend support and code generation}\label{backends} 
\head{Define-by-run graphs.}
While there are many advantages to defining models as end-to-end computation graphs, define-by-run semantics and eager execution are increasingly popular for ease of use. RLgraph's component graph can be executed both in static graph construction or define-by-run mode. To implement a PyTorch backend, we only had to modify the build procedure as follows. As there are no placeholders, we simply create torch tensors during the build phase as artificial placeholders to push through the dataflow graph for shape and type inference of variables (e.g. tensors used to store state). This is because even in define-by-run mode, automatically dealing with nesting and splitting of complex spaces across time and batch dimensions can be handled via decorators. Next, after building the component graph, we change the execution mode flag for API methods from 'build' to 'define-by-run'. In this mode, instead of returning operation objects used for graph construction, RLgraph simply directly evaluates a call-chain of graph functions to retrieve the result. RLgraph hence provides a unified interface for executing its component graph API in define-by-run and static-graph mode. 
 
\head{Autograph and graph optimization.}
An emerging but early trend in deep learning frameworks are features to automatically convert imperative code to static computation graphs. Examples of this include TensorFlow's AutoGraph \cite{Moldovan2018} mechanism and PyTorch's JIT tracing. We encountered temporary limitations e.g. in AutoGraph on stateful operations which would need to convert list manipulation to TF variable updates. We plan to merge backend-dependent graph-function implementations into single-stream functions, which will then be auto-converted where possible.

The ad-hoc mixed-backend implementation style of many RL libraries creates hurdles for systematically exploiting graph generation. RLgraph provides a natural fit for these features due to its organization of all backend code into graph functions as logical units. Features such as auto-graph can hence be integrated centrally in decorators during the build process (similar to device assignment).

RLgraph's separation of concerns opens up opportunities for optimization at all stages. Emerging approaches in optimizing execution (e.g. via automated device placement \cite{hierarchical2018}), or backend-dependent compilations can be integrated at the graph build stage. 

\section{Evaluation}\label{evaluation}
We evaluate RLgraph's performance using different distributed execution engines, local deep learning backends, and device strategies. Our aim is not to benchmark the underlying frameworks but to show RLgraph can perform competitively compared to native implementations.

\subsection{Results}

\head{Build overhead.}
We begin by comparing the one-time build overhead of RLgraph's abstractions on different backends. Recall there are two sources of overhead during initialization. First, the component graph is created by traversing the graph from the root. Second, creating variables and potentially static graph operations by letting input spaces flow through the component graph requires moving through components in the iterative build procedure.

\begin{figure}[ht]
\centering
\begin{subfigure}[t]{.235\textwidth}
\includegraphics[width=\textwidth]{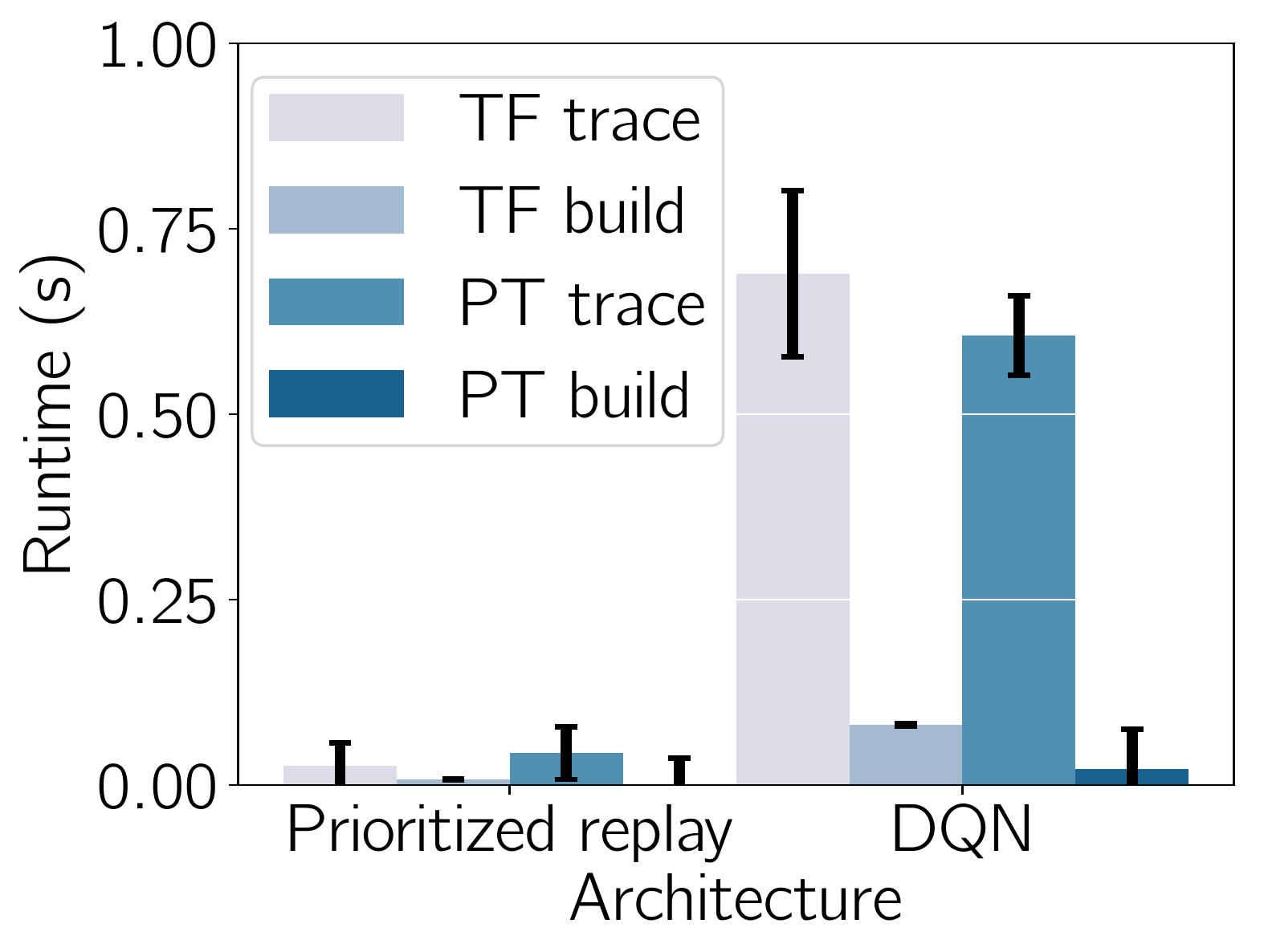}
\caption{\label{fig:build-ovheread} Build overheads.}
\end{subfigure}
\begin{subfigure}[t]{.235\textwidth}
\includegraphics[width=\textwidth]{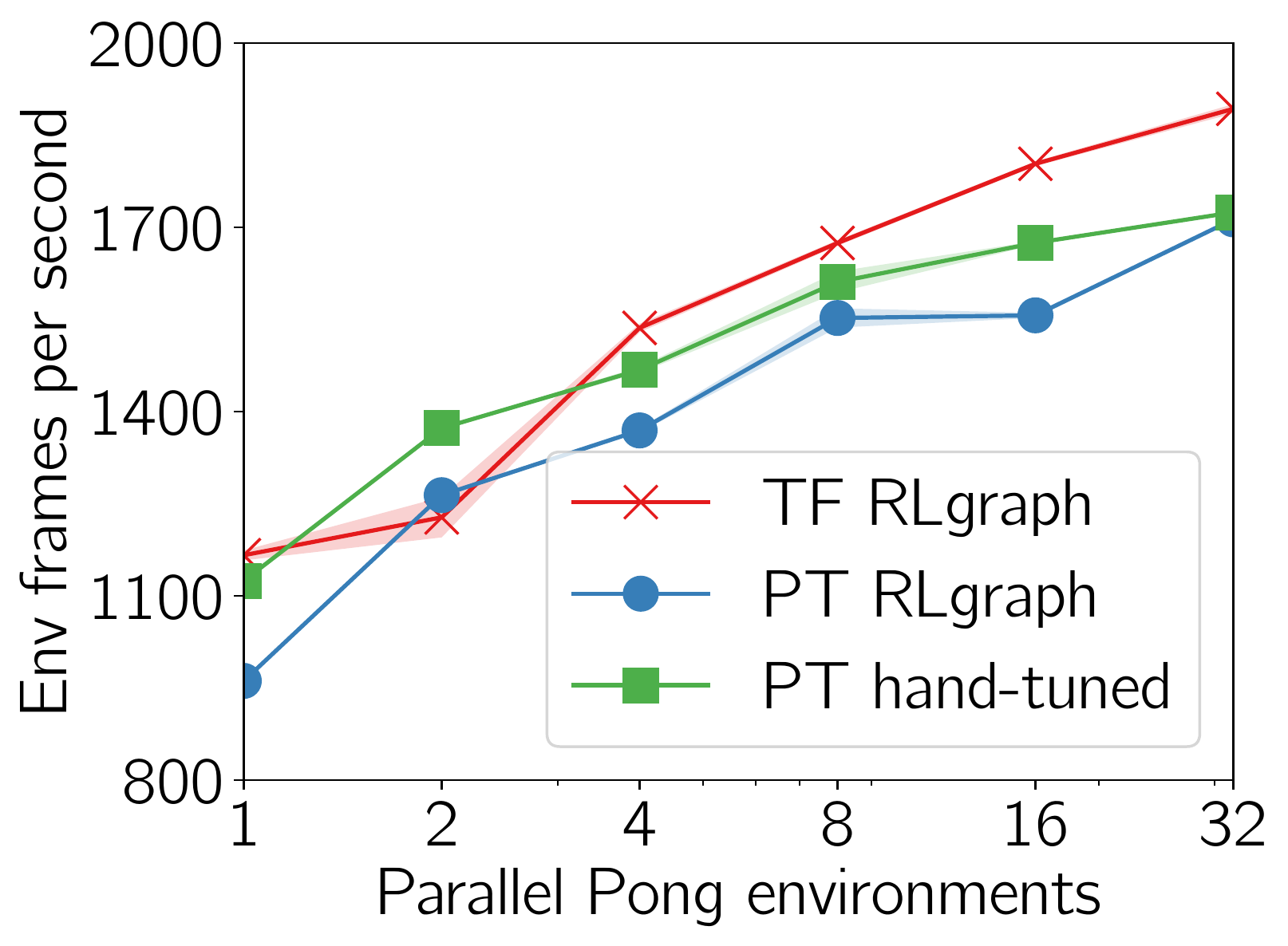}
\caption{\label{fig:act-performance} Worker act performance.}
\end{subfigure}
\caption{TensorFlow and PyTorch backend comparison.}
\vspace{-2mm}
\end{figure}

In Figure \ref{fig:build-ovheread}, we show component graph and main build overhead on a single memory component and a common RL architecture (DQN) for the TensorFlow (TF) (v1.11) and PyTorch (PT) (v0.4) backends.  Overhead here refers to the time spent on top of creating variables and operations, which would have to be done irrespective of using RLgraph. The overhead for both build phases to build a single component (e.g. for modular performance testing and debugging) is less than 100 ms. For a common architecture (dueling DQN with prioritized replay, 43 components), the combined overhead is about 1 s for TF and 650 ms for PT. PyTorch builds only require a few milliseconds as variables (e.g. to represent memory state) are native Python lists or NumPy arrays.

We also compare backend runtime performance by testing act (inference) throughput on a single-threaded worker acting on a vector of environments (Fig. \ref{fig:act-performance}). We use the Atari Pong environment and a standard 3-layer convolutional architecture followed by a dueling network for action selection. The TF backend (\textit{TF RLgraph}) does not incur runtime overhead because the component graph is discarded after building. The graph executor only looks up the name of the operation in the op registry and executes the corresponding TF op. In define-by-run mode using PyTorch (\textit{PT RLgraph}), RLgraph incurs some overhead when calls are routed through components. To understand this overhead, we also implemented a bare-bones PyTorch actor, including fine-tuning OpenMP and MKL settings (\textit{PT hand-tuned}). 

TensorFlow outperforms both PyTorch variants as batch-size increases on a CPU, thus making it more suitable for performing batch-acting and batch-postprocessing. RLgraph's PT backend carries overhead due to requiring additional lookups when traversing the graph via API decorators. This overhead becomes negligible as batch size increases and  runtime is dominated by the network forward passes. To reduce this overhead when traversing the component graph, we implemented some initial fast-path calls. For some cases, the graph builder can identify edge-contractions (where calls are edges and components are vertices), so define-by-run execution through the relevant sub-graph requires no intermediate component calls.

To the best of our knowledge, RLgraph represents the first common interface to TF and PT on a component level. High-level APIs like Keras only support static graph approaches. Libraries such as Ray RLlib support TF/PT backends but only on an algorithm level where entire implementations can be parallelized via RLlib's distributed abstractions. Using RLgraph, developers assemble logic via the backend-independent API, and can then implement e.g. a new loss component in their backend of choice.

\head{Distributed execution on Ray.}
Next, we evaluate RLgraph on the distributed execution engine Ray in comparison to Ray's native library, RLlib (v0.5.2). We implemented distributed prioritized experience replay (Ape-X \cite{Horgan2018}), a state-of-the-art distributed Q-learning algorithm on our Ray executor. We further implemented a vectorized environment worker for sample collection, including all heuristics described in the Ape-X paper and found in RLlib's implementation (e.g. worker-side prioritization).

All experiments were performed on Google Cloud with the learner being hosted on a GPU instance with 1 active V100 GPU, 24 vCPUs and 104 GiB RAM. Sample collection nodes had 64 vCPUs and 256 GiB RAM.
\begin{figure}[ht]
\centering
\includegraphics[scale=.3]{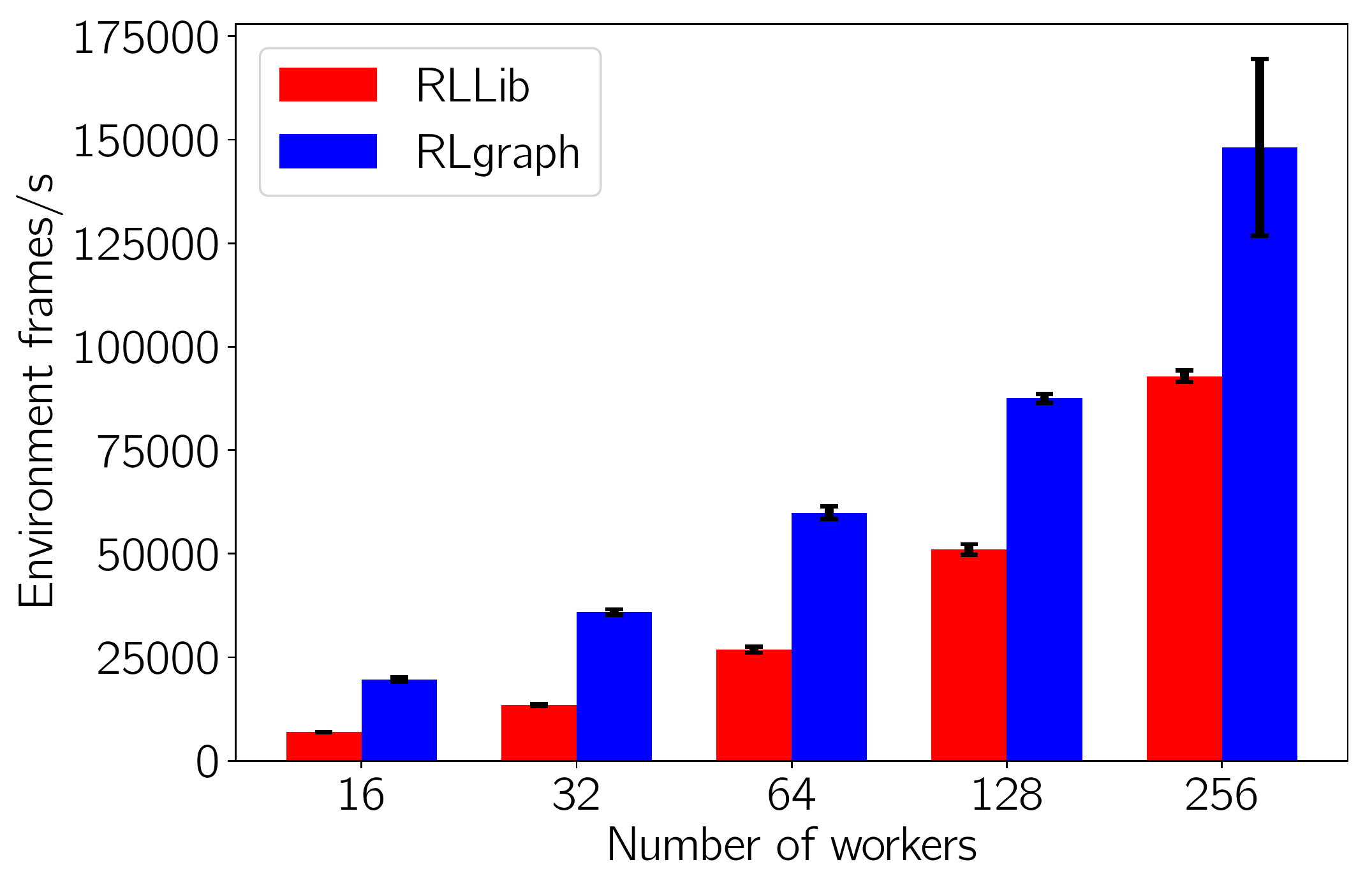}
\vspace{-2mm}
\caption{\label{fig:apex-tp} Distributed sample throughput on \textit{Pong}.}
\vspace{-2mm}
\end{figure}
Figure \ref{fig:apex-tp} shows sampling performance on the Pong environment. The x-axis represents the number of policy-evaluators/Ray-workers respectively (RLlib, RLgraph), each initialized with a single CPU, and the y-axis shows environment frames per second (including frame skips). Each worker executed 4 environments, and we used 4 instances of replay memories to feed the learner (we did not observe improvements using more memories due to limited vCPUs on GPU nodes). All settings were run with 8 sample nodes except 256 workers (16 sample nodes) to ensure sufficient memory. RLgraph outperforms RLlib by a large margin (185\% on 16, 60\% on 256 workers) despite implementing the same algorithm with equivalent hyper-parameters, model size, and environments. Performance for 16 workers is highest due to better resource utilization. RLgraph also completes its learning tasks faster due to improved implementations.

The reason for RLgraph's performance is systematic component analysis yielding insights into efficient sample processing. For example, RLlib's policy evaluators execute multiple session calls to incrementally post-process batches. RLgraph instead splits post-processing in incremental and batched parts to minimize calls to the TensorFlow runtime.
\begin{figure}[ht]
\centering
\begin{subfigure}[t]{.235\textwidth}
\includegraphics[width=\textwidth]{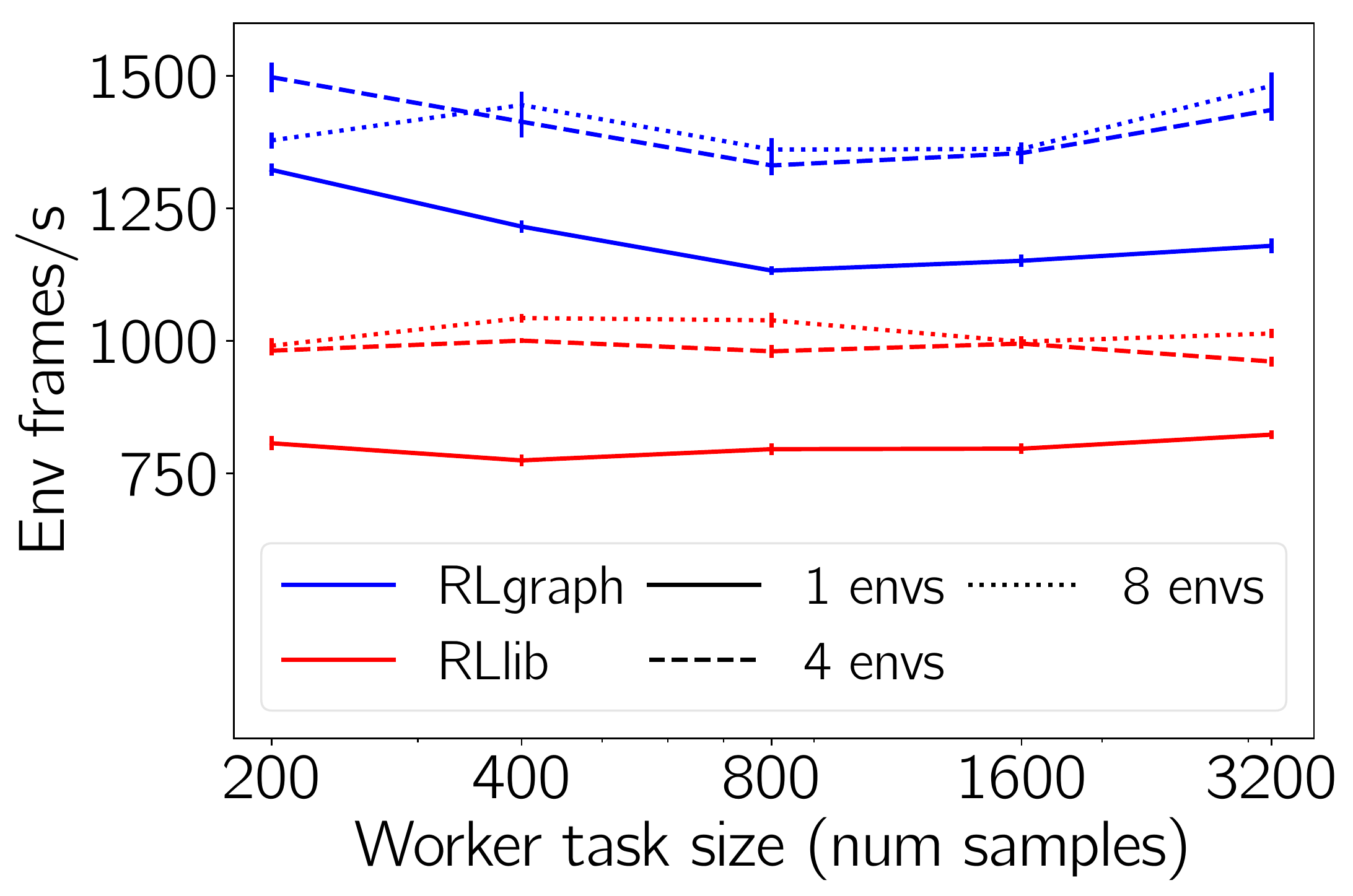}
\caption{\label{fig:apex-worker} Single worker throughput.}
\end{subfigure}
\begin{subfigure}[t]{.235\textwidth}
\includegraphics[width=\textwidth]{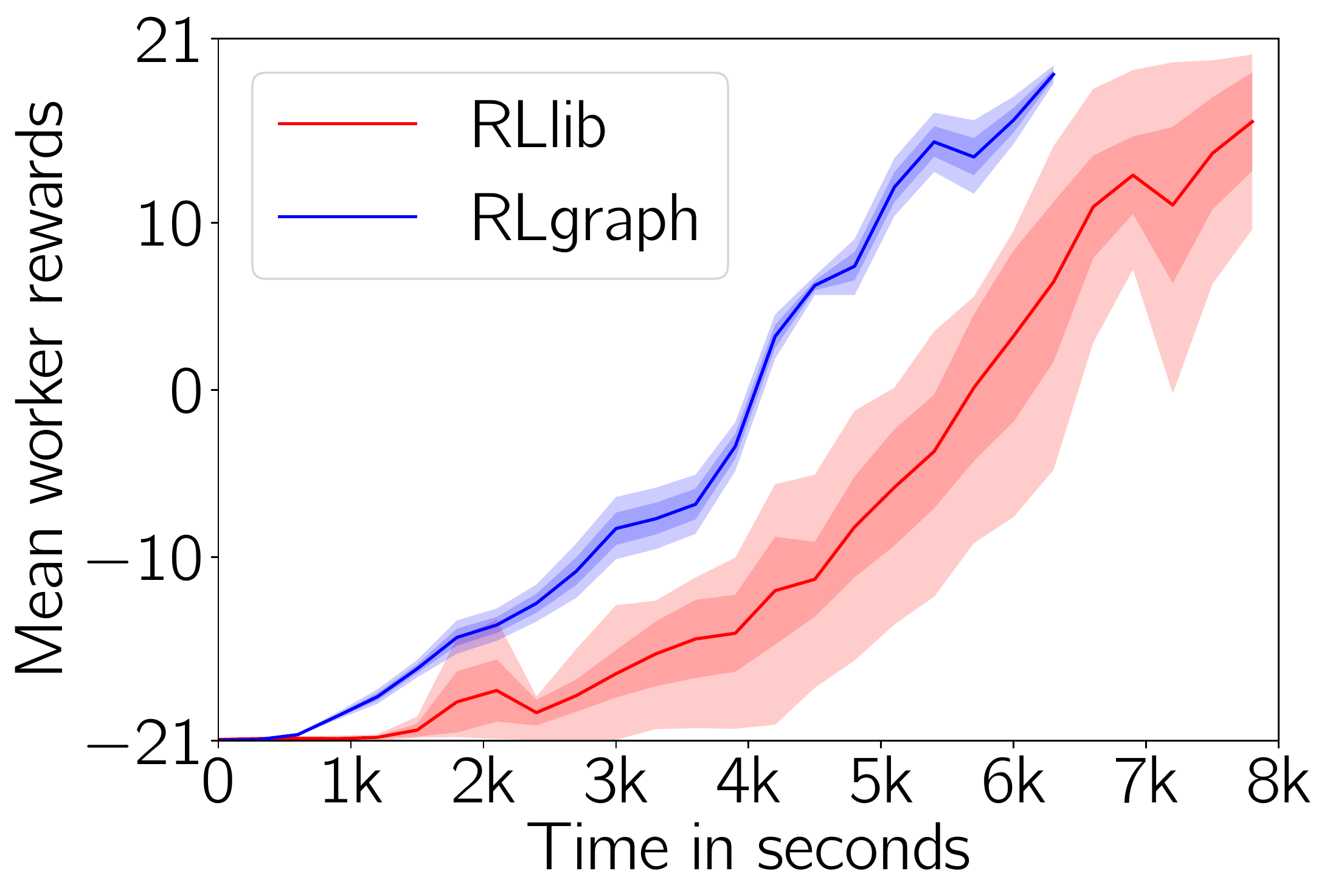}
\caption{\label{fig:apex-learning} Training times for \textit{Pong}.}
\end{subfigure}
\caption{Single task throughput and learning comparison.}
\vspace{-2mm}
\end{figure}
These effects can also be observed at the scale of a single task across different task lengths and number of environments (called sequentially). Figure \ref{fig:apex-worker} shows the requested number of samples versus the achieved frames per second using a single RayWorker (RLgraph) versus a policy evaluator (RLlib). Both use the same agent and configuration as in the distributed setting (10 warm up runs, mean across 50 runs). RLgraph is not only more effective on a single environment, it also scales better on vectorized environments due to faster accounting across environments and episodes.

We show learning results (Fig. \ref{fig:apex-learning}) to confirm that RLgraph's throughput is not at the cost of training performance. We use RLlib's provided tuned \textit{Pong} configuration (32 workers). In RL, the same code using the same hyper-parameters can vary drastically across runs, so reliably comparing learning is difficult \cite{Henderson2017}. We ran 10 random seeds and averaged across the 3 best runs (both libraries did not learn anything for some seeds as expected). In line with throughput, RLgraph learns to solve (reward 21) Pong substantially faster than RLlib.

Note that RLlib's published results on Ape-X throughput do not include updating without stating this explicitly, and later reported results including updates\footnote{Source: RLlib authors, \url{https://github.com/ray-project/ray/issues/2466}}  are up to 130 k frames per second on 256 workers (versus 170 k max for RLgraph). Experimental versions of Ray's new backend include improved garbage collection of which RLgraph would benefit to the same extent as RLlib\footnote{During experimentation using Ray's original backend, we experienced some difficulties with the Ray engine, including memory leaks and task initialization crashes.}. Our results show RLgraph's execution-agnostic design can integrate with external execution engines, and perform competitively. While RLlib could adopt more efficient implementations, our insight is that RLgraph can be used on Ray via few wrapper classes. Implementing other distributed semantics on Ray with RLgraph only requires extending the generic Ray executor to implement a coordination loop. Finally, RLgraph's modularization helps visualizing computation graphs when compared to programming models such as RLlib's, which scatter code across fragmented tasks. We visualize both Ape-X implementations in Appendix \ref{apex-vizualisations}.
\begin{figure}[h]
\centering
\includegraphics[scale=.25]{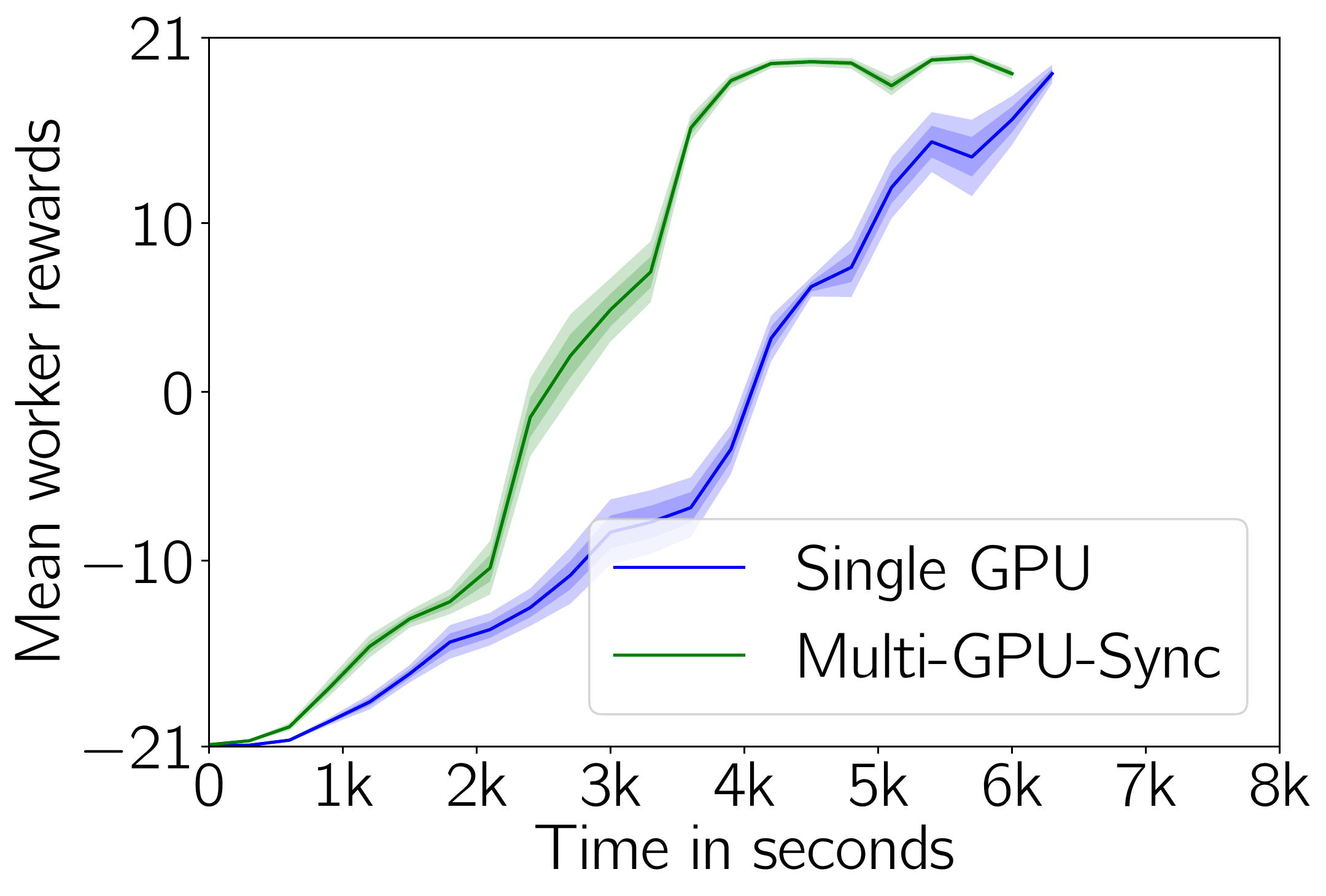}
\vspace{-2mm}
\caption{RLgraph synchronous multi-GPU device strategy \label{fig:multi-gpu-learning}}
\vspace{-4mm}
\end{figure}

\head{Multi-GPU strategies.} We also implemented device strategy prototypes where the component graph is automatically expanded during the build phase (e.g. to create sub-graph replicas). When using a synchronous GPU replica strategy, the update batch is internally split into multiple sub-batches, and gradients are averaged across towers. Fig. \ref{fig:multi-gpu-learning} contains Ape-X results using 1 and 2 V100 GPUs. We observe the expected speed-up in convergence.

\head{Distributed TensorFlow.} Finally, we evaluate RLgraph using the distributed TensorFlow backend on DeepMind's (DM) importance-weighted actor-learner architecture (IMPALA) \cite{Espeholt2018}. The authors have open-sourced an optimized implementation\footnote{Code at: \url{https://github.com/deepmind/scalable_agent}}. IMPALA perhaps best represents the end-to-end computation graph paradigm, where even environment interaction is fused into the TF graph. We implemented IMPALA in RLgraph to demonstrate its ability to generate such graphs. To this end, RLgraph provides generic execution components for graph-fused environment stepping. IMPALA executes updates by letting each actor perform a rollout step (100 samples) and input its samples into a globally shared blocking queue. The learner dequeues rollouts and uses a staging area to hide GPU latency. 
\begin{figure}[h]
\centering
\includegraphics[scale=.3]{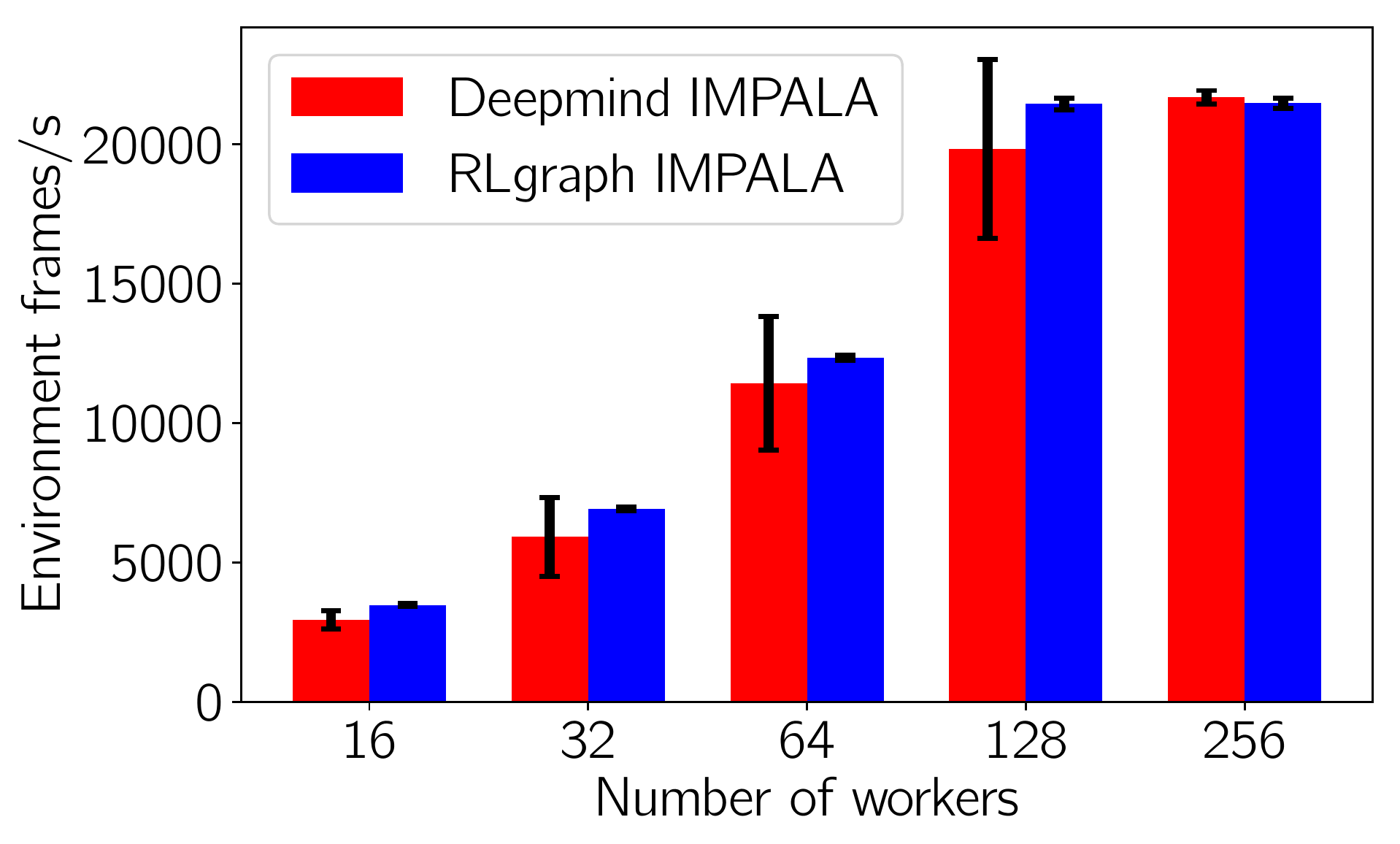}
\vspace{-2mm}
\caption{IMPALA throughput comparison on \textit{seekavoid\_arena\_01} \label{fig:impala-throughput}}
\vspace{-5mm}
\end{figure}
Figure \ref{fig:impala-throughput} compares throughput using the large network described in the paper on a DM lab 3D task (which are more expensive to render than Atari tasks). We again use a single V100 GPU for the learner and let 4 workers each share a 8 vCPU instance. RLgraph achieves about 10-15\% higher mean throughput (5 runs) for fewer workers until both implementations are limited by updates. DM's implementation exhibited higher variance due to subtle differences in preprocessing tensors after unstaging. DM's code also carried out unneeded variable assignments in the actor. Removing these yielded 20\% improvement in a single-worker setting for RLgraph. Emerging tools in graph optimization (e.g. TF XLA, TVM \cite{Chen2018}) can help with optimizing data layouts and numerical transformations. RLgraph helps most in improving high level dataflow as it enforces well-defined interactions between components (c.f. Appendix \ref{impala-vizualisations}). This in turn improves reasoning about the complex design patterns found in RL algorithms.

\subsection{Discussion}
Our results show that using RLgraph to define algorithms via combining backend-independent components yields high-performing implementations across backends. Assembling the graph through the different build phases only adds up to one second of build overhead. Using Ray as a distributed backend, we demonstrated that RLgraph run on its own Ray executor outperforms Ray RLlib. This indicates that wide-spread mixed backend (e.g. Python/TF) RL implementations can be significantly accelerated via careful dataflow analysis, and without changing algorithm logic. Using distributed TF, we found that RLgraph can help improve dataflow in end-to-end static computation graphs. Our main take-away is that developers can use RLgraph to focus on logical component composition in RL independently of the underlying execution paradigm.

\section{Related work}
\label{related}
RLgraph builds upon the experiences of many prior libraries which we discussed in \S \ref{existing-libraries}. Here, we discuss emerging trends in programming models and optimizations.

\subsection{Programming models}
The success of deep learning frameworks has given rise to several higher-level learning APIs seeking to free users from dealing with lower level tensor operations. Keras \cite{chollet2015keras} is a popular framework for quick assembly and training of deep learning models with support for multiple static-graph backends (e.g. TensorFlow \cite{abadi2016tensorflow}, CNTK \cite{seide2016cntk}, MXNet \cite{chen2015mxnet}). Gluon provides a concise API for imperative, dynamic neural networks on top of MXNet \cite{gluon}. TF.Learn offers a high level API for constructing symbolic TensorFlow graphs \cite{Tang2016}. Among these frameworks, TensorFlow's programming model \cite{Abadi2017} is distinct because it supports in-graph control-flow \cite{Yu2018}. A key issue in the high-level APIs above is that they typically assume control-flow to be implemented in the driver language (frequently Python). RLgraph bridges this gap by providing Sonnet-style \cite{sonnet} composable components, with the addition of API methods handling in-graph control flow, or define-by-run graph execution.

\subsection{Graph optimizations and code generation}
Programming models for machine learning and in particular deep learning typically prioritize high level APIs and usability over performance. Code is written in a multitude of languages and libraries, and is executed over a variety of backends. Frameworks like Weld \cite{Palkar2017} optimize performance by integrating library calls into a common intermediate representation which is then mapped to efficient multi-threaded code.
Mirhoseini et al. identify effective TensorFlow device placements via hierarchical reinforcement learning \cite{hierarchical2018}. FlexFlow further improves parallelization strategies for specific deployments using an execution simulator and fine-grained randomized search across execution dimensions \cite{Jia2018}. TVM is a compiler stack to optimize tensor operations across diverse hardware backends \cite{Chen2018}.  RLgraph constructs component graphs irrespective of execution semantics. Optimizations can be performed at the level of graph executors by including them in the build.

The proliferation of deep learning frameworks which often focus development efforts on Python front-ends has also created the need for new shared representations. The aim of standard formats such as ONNX \cite{onnx} is to enable model interoperability, define a common route from prototyping to production deployment, and to create a shared runtime for optimizations. From a RL perspective, the aim of deployable graphs is more difficult to achieve due to the extensive control-flow and state management in RL workloads. Implementing code using e.g. TensorFlow control flow operators allows immediate export, optimization and deployment of the entire RL program but can slow down development. Framework developers have recognized this tension and are designing tools such as AutoGraph in TF for automatic graph generation. As these initiatives mature, we expect RLgraph components to merge backend-dependent graph functions by leveraging automatic graph generation.

\section{Conclusion}
\label{conclusion}
RLgraph is a new open source framework for designing and executing computation graphs for reinforcement learning. RLgraph's component graph abstraction allows developers to separate the composition of logical components in a RL algorithm from their local backend and distributed execution. The resulting implementations are fast, robust, incrementally testable, and easy to extend or re-use.

\section*{Acknowledgments}
Michael Schaarschmidt is supported by a Google PhD Fellowship. We are also grateful for receiving research credits from Google Cloud. This research was supported by the EPSRC (grant references EP/M508007/1 and EP/P004024), the Alan Turing Institute, and a Sansom scholarship. Further, we thank Lasse Espeholt for providing support to replicate IMPALA results. We also want to thank the RLlib authors for helping to replicate Ape-X results.

\bibliographystyle{sysml2019}

\newpage
\appendix
\section{Visualizing computation graphs}
\subsection{IMPALA}\label{impala-vizualisations}
To illustrate how RLgraph helps organize computation graphs, we visualize the IMPALA implementation discussed in the evaluation. 

\begin{figure}[h]
\centering
\includegraphics[scale=.4]{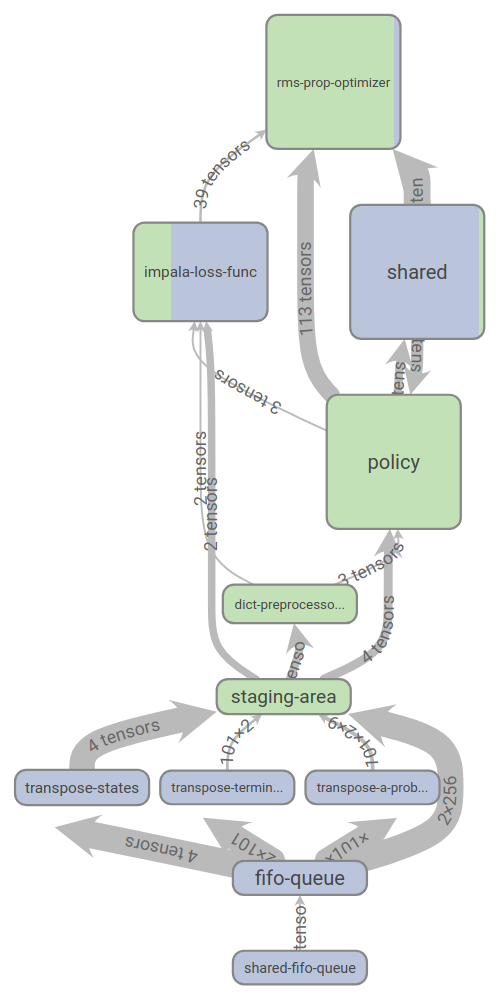}
\caption{TensorBoard visualization of RLgraph's IMPALA learner. As all operations and variables are organized in components under separate scopes, dataflow between components is clear. \label{fig:impala-learner-rlgraph}}
\end{figure}

Figure \ref{fig:impala-learner-rlgraph} illustrates the RLgraph implementation using TensorBoard. As each component's scope and variables are managed by RLgraph during the build, device assignments and dataflow are easy to visualize automatically with existing tools. Here, green components are on the GPU while blue components are on the CPU. The dataflow from bottom to top clearly illustrates how tensors are moved from the shared queue, preprocessed, then moved to a staging area. The prior batch is taken from the staging area, then preprocessed and passed to the policy and loss function. The optimizer and policy interact with a \textit{shared} scope as learner and workers share policy variables.

Mixed colours imply that a component has multiple sub-components on different devices. For example, the IMPALA loss-function computes an importance correction on the CPU as it is difficult to parallelize, so the loss-component has mixed assignments. We found that RLgraph specifically helps to make more effective use of existing tools like TensorBoard, as clean visualizations are key to identifying problems (e.g. with device assignments).

\begin{figure*}[t]
\centering
\includegraphics[scale=.28]{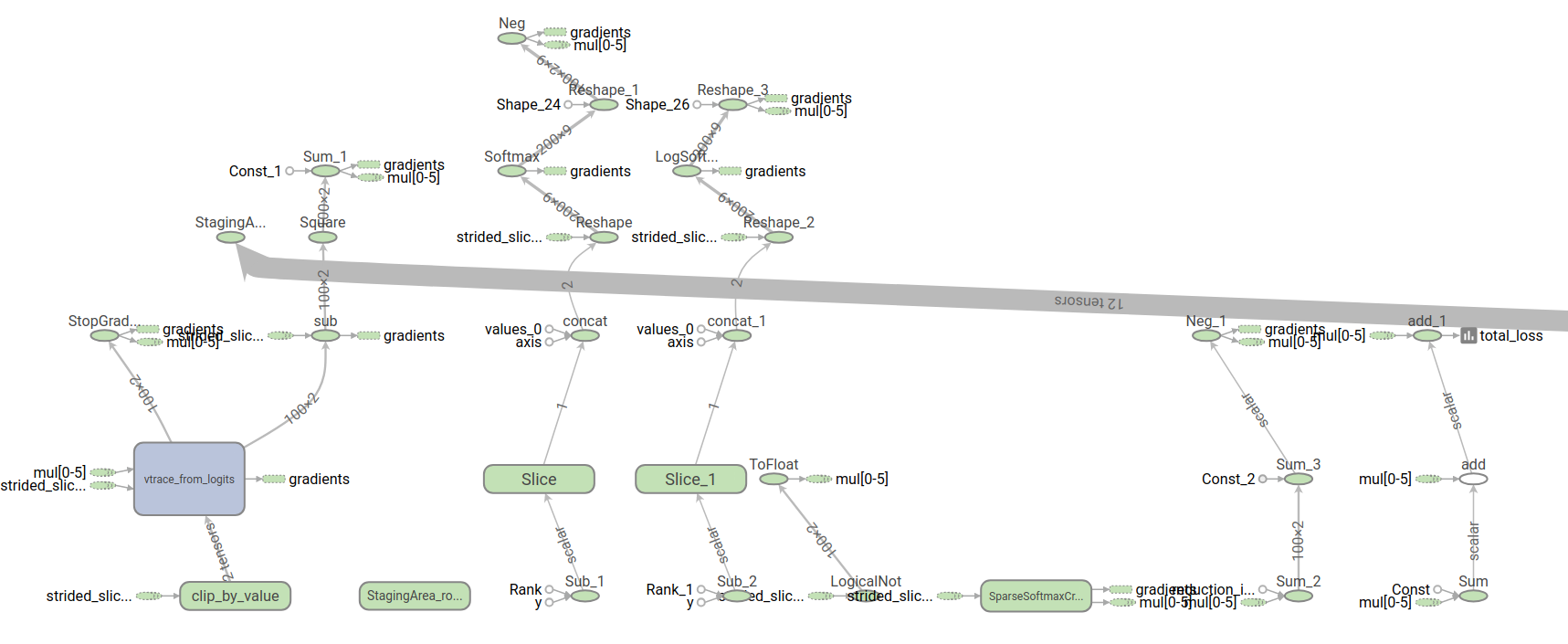}
\caption{TensorBoard visualization of DeepMind's IMPALA learner (left). \label{fig:impala-learner-dm-left}}
\end{figure*}

\begin{figure*}[t]
\centering
\includegraphics[scale=.28]{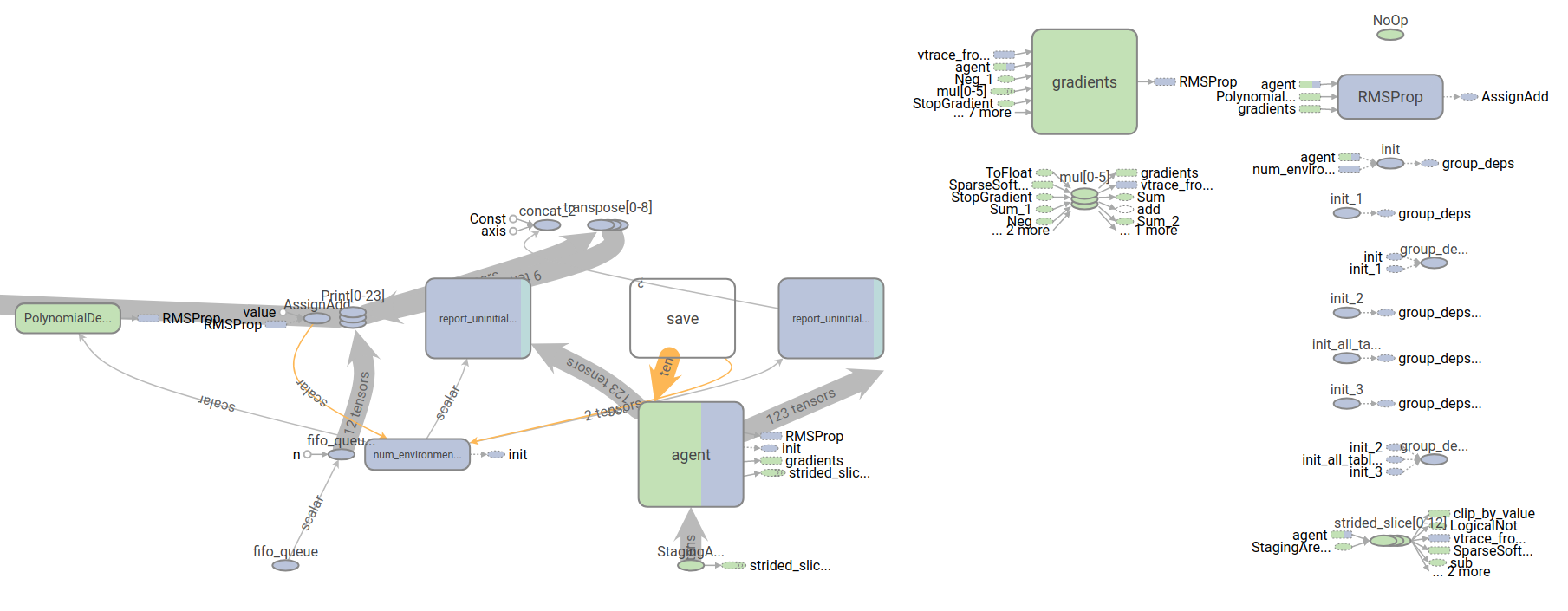}
\caption{TensorBoard visualization of DeepMind's IMPALA learner (right). \label{fig:impala-learner-dm-right}}
\end{figure*}

In Figures \ref{fig:impala-learner-dm-left} and \ref{fig:impala-learner-dm-right} (split due to size in TensorBoard), we show for comparison a visualization of DeepMind's open source IMPALA implementation. As is common for self-contained RL scripts, scopes, devices, and operation names are handled on an ad-hoc basis. The resulting graph, while implementing the same logic as our code, is highly fragmented in the visualization, making insights into potential problems much more difficult.


\subsection{Ape-X}\label{apex-vizualisations}
\begin{figure}[t]
\centering
\includegraphics[scale=.5]{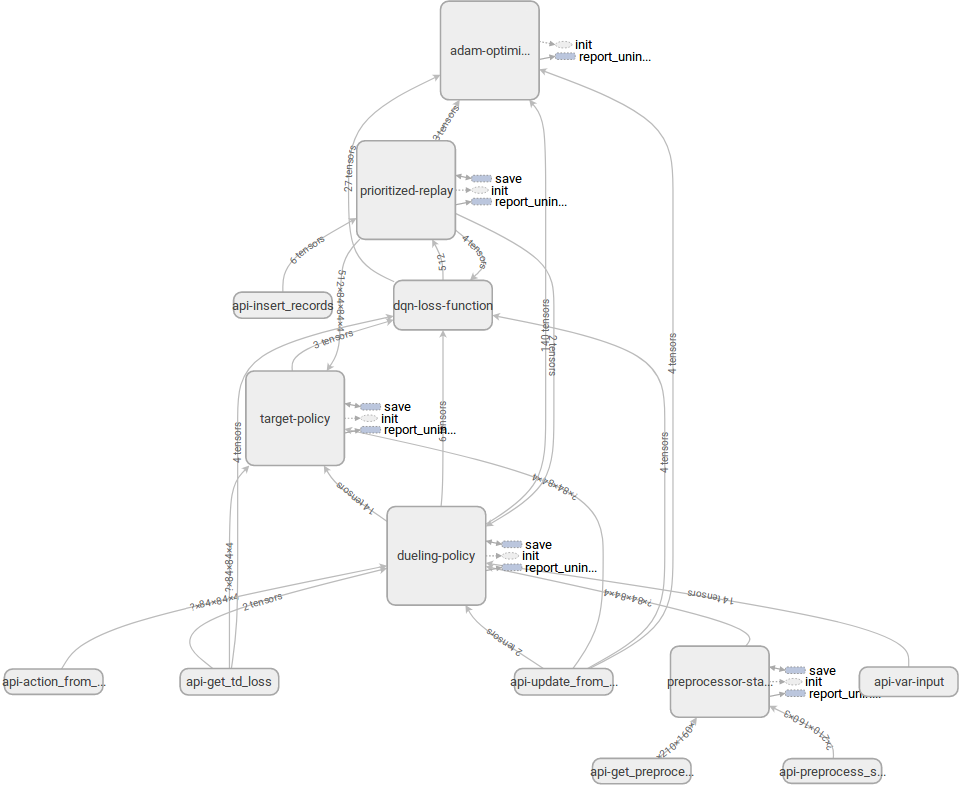}
\caption{TensorBoard visualization of RLgraph's Ape-X learner. \label{fig:apex-learner-rlgraph}}
\end{figure}

\begin{figure*}[h]
\centering
\includegraphics[scale=.28]{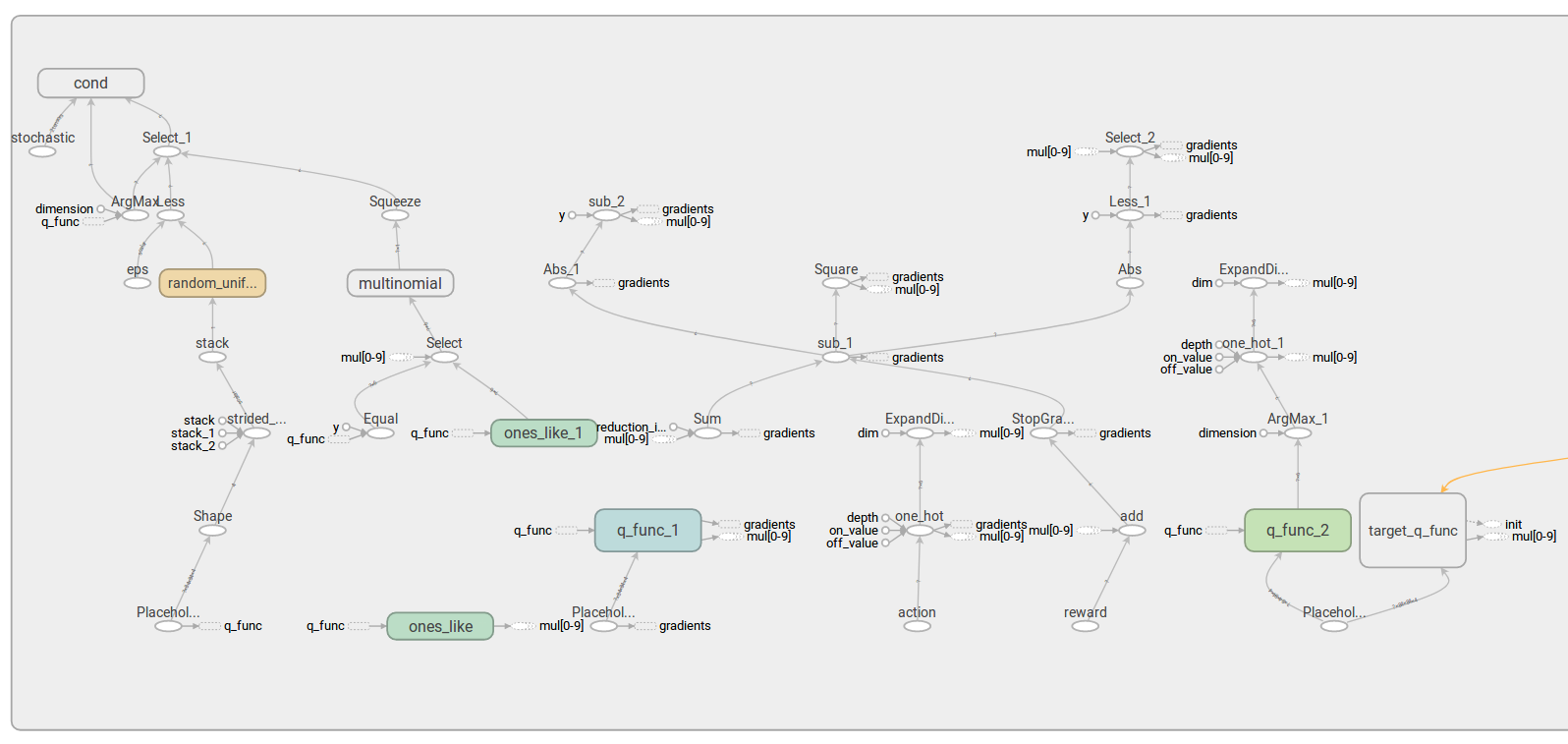}
\caption{TensorBoard visualization of RLlib's Ape-X learner (left). \label{fig:apex-llib1}}
\end{figure*}

\begin{figure*}[h]
\centering
\includegraphics[scale=.28]{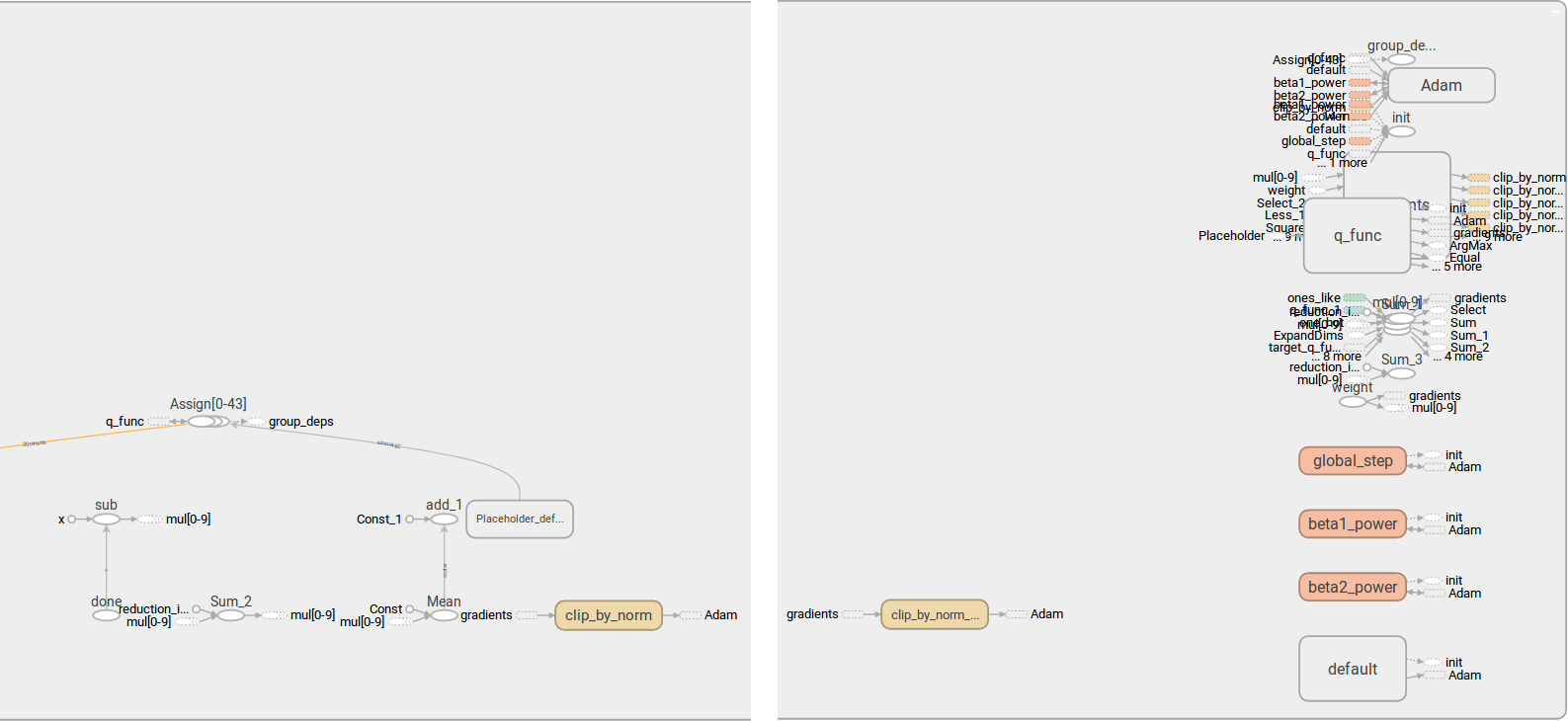}
\caption{TensorBoard visualization of RLlib's Ape-X learner (right). \label{fig:apex-llib2-3}}
\end{figure*}

When executing on Ray, computation graphs are fragmented into separate tasks represented by Ray actors, as opposed to end-to-end differentiable graphs such as in distributed TensorFlow \cite{Abadi2017,Yu2018}. In RLlib, task code is scattered across agent classes, policy graphs, and backend-specific utilities. Understanding dataflow between components is difficult due to a lack of consistent modularization across imperative function calls. RLgraph's separation of execution semantics and graph design results in consistent graph creation irrespective of the backend used. In Figure \ref{fig:apex-learner-rlgraph}, we show the corresponding TensorBoard visualization for RLgraph's Ape-X learner. Figures \ref{fig:apex-llib1} and  \ref{fig:apex-llib2-3} show RLlib's highly fragmented Ape-X graph.

\end{document}